\documentclass{article}

\usepackage[legalpaper, margin=1in]{geometry}
\usepackage{graphicx} % Required for inserting images
\usepackage{hyperref}
\usepackage{chngcntr}
\usepackage{natbib}
\usepackage{booktabs}
\usepackage{tabularx}
\usepackage{multirow}
\usepackage{mdframed}
\usepackage{pifont}
\usepackage[normalem]{ulem} % strikethrough
\usepackage{CJKutf8} 

\usepackage{footnote}
\makesavenoteenv{tabular}
\makesavenoteenv{table}

\usepackage{siunitx}
\usepackage{makecell}

\counterwithin*{section}{part}

\title{Advancing the Evaluation of Traditional Chinese Language Models:
Towards a Comprehensive Benchmark Suite}
\author{\small Chan-Jan Hsu\thanks{These authors contributed equally to this work and are arranged in alphabetical order} , Chang-Le Liu$^*$, Feng-Ting Liao$^*$, Po-Chun Hsu$^*$, Yi-Chang Chen$^*$, Da-shan Shiu \\ MediaTek Research}
\date{September 2023}

\begin{document}
% \part{Public release version}
\maketitle

\begin{abstract}
The evaluation of large language models is an essential task in the field of language understanding and generation. As language models continue to advance, the need for effective benchmarks to assess their performance has become imperative. In the context of Traditional Chinese, there is a scarcity of comprehensive and diverse benchmarks to evaluate the capabilities of language models, despite the existence of certain benchmarks such as DRCD, TTQA, CMDQA, and FGC dataset. To address this gap, we propose a novel set of benchmarks that leverage existing English datasets and are tailored to evaluate language models in Traditional Chinese. These benchmarks encompass a wide range of tasks, including contextual question-answering, summarization, classification, and table understanding. The proposed benchmarks offer a comprehensive evaluation framework, enabling the assessment of language models' capabilities across different tasks. In this paper, we evaluate the performance of GPT-3.5, Taiwan-LLaMa-v1.0, and Model 7-C, our proprietary model series, on these benchmarks. The evaluation results highlight that Model 7-C achieves performance comparable to GPT-3.5 with respect to a part of the evaluated capabilities. In an effort to advance the evaluation of language models in Traditional Chinese and stimulate further research in this field, we have open-sourced our benchmark and opened the model for trial.

\end{abstract}

\section{Introduction}
The evaluation of large language models has long been a crucial task. With the advancement of technology, language models have become more sophisticated, providing higher-quality responses akin to human responses to open-ended questions. However, evaluating these models is challenging, and there is a need for well-designed benchmarks to assess their performance comprehensively and consistently. Existing English benchmarks such as MMLU \citep{hendryckstest2021}, IMDB \citep{maas-EtAl:2011:ACL-HLT2011}, and XSum \citep{Narayan2018DontGM} cover measurements of models' capabilities in question answering, sentiment classification, and summarization, respectively. In Traditional Chinese, while there exist some benchmarks such as Delta Reading Comprehension Dataset (DRCD)  \citep{shao2019drcd}, Taiwanese Trivia Question Answering (TTQA) \citep{ennen2023extending}, and Formosa Grand Challenge (FGC) dataset \citep{FGC}, there is limited availability of comprehensive and diverse benchmarks for evaluating language models' capabilities. \\

In this paper, to address the need for a comprehensive suite of evaluations in Traditional Chinese, we propose a set of new benchmarks. The benchmarks are built upon available Traditional Chinese and English datasets to test the capabilities of language models in Traditional Chinese. Our proposed benchmarks assess the capabilities of tasks related to contextual question answering, world knowledge, summarization, classification, table understanding. In terms of evaluating world knowledge, we further propose a new dataset - Taiwan Massive Multitask Language Understanding (TMMLU) - encompassing exams from high school entrance exams to vocational exams across 55 subjects in total. \\
%As to evaluation of Traditional Chinese models on open-ended questions, we leverage TAIDE-14 \citep{taide14_2023}, a suite of questions across 14 tasks. We employ GPT-4 as a judge to quantify the helpfulness performance of models in Traditional Chinese. \\

We evaluate the performance of proprietary and open-source models, namely GPT-3.5, Taiwan-LLaMa-v1.0 \citep{lin-chen-2023-llm}, Model 7-C (ours) and Model 7-C-Chat (the fine-tuned version of Model 7-C for chatting capability), using our proposed Traditional Chinese benchmarks. Notably, our proposed benchmarks provide a comprehensive set of evaluation tasks for language models, allowing us to assess their performance on various tasks. For some of the evaluated capabilities, the evaluation outcomes demonstrate that Model 7-C matches the performance of the state-of-the-art GPT-3.5 model in Traditional Chinese. \\

To promote more research on advancing state-of-the-art language models in Traditional Chinese, we have open-sourced our benchmark code and relevant datasets and opened for trial of our proprietary model, Model 7-C, for comparison\footnote{\url{https://github.com/mtkresearch/MR-Models}}.

% Please write a proper introduction.
% (I moved some inappropriate paragraphs from Section II over here.) 
% At the time of this writing (early August, 2023), the current benchmarks in Traditional Chinese inadequately represent the full scope of tasks that LLM can perform. 
% In summary, the four mentioned Traditional Chinese datasets focus solely on the contextual question-answering task, whereas benchmarks like HELM evaluate language models across more than 16 tasks in English.

% \section{Traditional Chinese evaluations at August 2023}
\section{Related work}
\label{TC_evaluations_at_Aug23}
\sloppy
There exists a wealth of English benchmarks for evaluating different capabilities of language models. EluetherAI's Language Model Evaluation Harness \citep{eval-harness} is a unified framework to test generative language models on a large number of different evaluation tasks. Holistic Evaluation of Language Models (HELM) \citep{liang2022holistic} is an evaluation framework that consists of evaluations in 42 scenarios. BIG-bench \citep{srivastava2023beyond} is a collaborative benchmark designed to examine large language models across diverse task topics ranging from linguistics and childhood development to software development and social bias. AGIEval \citep{zhong2023agieval} is a benchmark tailored to assess models on human cognition and problem-solving, derived from 20 prominent admission and qualification exams including the Gaokao, SAT, law school tests, and civil service exams. These English benchmarks and the evaluations therein are commonly evaluated at the release of the models such as BLOOM \citep{scao2022bloom}, Pythia \citep{biderman2023pythia}, Falcon \citep{penedo2023refinedweb}, Llama (1 \citep{touvron2023llama} and 2 \citep{touvron2023llama2}), and their fine-tuned variants.\\

As to the notable open benchmarks in Traditional Chinese, at the time of this writing (mid-August, 2023), we summarize them below. DRCD, a reading comprehension peer-reviewed dataset, contains 30k question-answer pairs based on Wikipedia articles. TTQA \citep{ennen2023extending}, a trivia question-answering not-peer-reviewed dataset, consists of 64 expert-selected paragraphs from Wikipedia for testing a model's knowledge on Taiwanese-specific topics. Chinese Movie Dialogue Question Answering (CMDQA) \citep{luo-etal-2022-chinese}, a dialogue-based information-seeking question-answering dataset, contains 10k QA dialogues (40k turns in total) about movie information parsed from Wikipedia. Formosa Grand Challenge (FGC) dataset is a passage question answering dataset of 750 samples created from Taiwanese news articles and government announcements. \\

Language models have been shown to provide responses akin to human responses to open-ended questions. The open-ended types of questions however cannot easily be mapped 1-on-1 to a single answer. At the time of this writing, notable evaluation benchmarks with GPT-4 as judge have been wildly adopted by the community, albeit its tendency to favour longer text and texts generated by LLM \citep{lin-chen-2023-llm, liu2023geval}. Vicuna \citep{vicuna2023} consists of 80 questions spanning across 8 tasks. Similar to Vicuna, WizardLM \citep{xu2023wizardlm} constructed a test set of 218 open-ended questions covering 29 areas such as writing, role-play, and philosophy. As for Traditional Chinese open-ended questions, a translated version of Vicuna benchmark is used to test Taiwan-LLaMa \citep{lin-chen-2023-llm,taiwanllama}. In this study, we focus on benchmarks where ground truths are readily available.

% \section{Newly added Traditional Chinese evaluations}
\section{Benchmark}

Here we give a succinct introduction to each benchmark we will use in this study. We categorize the proposed set of benchmarks into capabilities. \autoref{measured_tasks_list} lists the evaluation benchmarks used in this study and \autoref{question_example} shows some examples. As source datasets in Traditional Chinese are limited, we translated the listed English datasets to Traditional Chinese for the evaluation.
%For helpfulness evaluation of models on open-ended questions, we use TAIDE-14 \cite{taide14_2023}, consisting of 14 prevalent generation tasks. 
% Furthermore, we carefully selected 280 prompts consisting of both single turn and multi-turn prompts from various sources, such as WizardLM, Chinese-LLaMA-Alpaca \citep{chinese-llama-alpaca} and also our own collection, for conducting helpfulness evaluations across models.

\begin{table*}[ht]
    \centering
    \begin{tabular}{|c|c|c|}

    % \begin{tabularx}{\textwidth}{|c|X|p{3cm}|}
    \hline
        \textbf{Capabilities} & \textbf{Evaluation Dataset} & \textbf{Source Language} \\
        \hline
        \multirow{2}{*}{Contextual QA} & DRCD \citep{shao2019drcd} & Traditional Chinese \\
        & FGC \citep{FGC} & Traditional Chinese \\ \hline
        \multirow{2}{*}{World Knowledge} & TTQA \citep{ennen2023extending} & Traditional Chinese \\
        % & AICUP-Truth \cite{} & TC \\
        & TMMLU (ours) & Traditional Chinese \\ \hline % -TC \citep{hendryckstest2021}
        \multirow{1}{*}{Summarization} & XSum-TC  \citep{Narayan2018DontGM} & English\\
        % & ECTSum-TC  \citep{mukherjee-etal-2022-ectsum} & English \\
        %& Scientific Papers-TC \cite{} & En\\ 
        \hline
        \multirow{1}{*}{Classification} & IMDB-TC \citep{maas-EtAl:2011:ACL-HLT2011} & English \\
        % & RAFT \cite{} & En \\
        % & Amazon Polarity-TC \citep{NIPS2015_250cf8b5} & English \\ \hline
        % & Yelp Review \cite{} & En \\ 
        \hline
        %Reasoning & GSM8K-TC \citep{cobbe2021gsm8k} & English \\ \hline
        %Rewriting  & ParaSCI-TC \cite{} & English \\ \hline
        %Output Format Generation & Amazon Reviews Multi-TC \cite{} & English \\ \hline
        Table Understanding & Penguins-in-a-Table-TC \citep{srivastava2023beyond} & English \\ \hline
        % Role-play & OASST1 \cite{} & En \\ \hline
        % Literature-writing & Alpaca \cite{} & En \\ \hline
        % Tool Using & ToolQA-TC \citep{zhuang2023toolqa} & English \\ \hline
    % \end{tabularx}
    \end{tabular}
    \caption{The datasets and their respective nature for benchmarking capabilities in this study. We translated English datasets to Traditional Chinese for the evaluation, which is indicated by the ``-TC'' suffix.}
    \label{measured_tasks_list}
\end{table*}

\subsection{Capabilities}
Below are summaries of the benchmarked capabilities as listed in Table \ref{measured_tasks_list} and the corresponding datasets used in evaluating the respective capabilities in this study.\\

\textbf{Contextual Question Answering} is the task in which a model is given a contextual input and is asked to respond to a given question related to the input. This task is most similar to standard benchmarks in closed QA or common sense reasoning. DRCD is a Traditional Chinese machine reading comprehension dataset containing 10,014 paragraphs from 2,108 Wikipedia articles and over 30,000 questions. FGC dataset is a passage question answering dataset of 750 samples created from Taiwanese news articles and government announcements.\\
%, intended as a standard source for transfer learning.

\textbf{World Knowledge} task requires a model to have a certain level of knowledge about the real world.
TTQA is for assessing language models' common sense abilities on Taiwanese terms, comprising 64 passages from Wikipedia about diverse Taiwanese cultural topics, necessitating model comprehension and reasoning. Taiwan Massive Multitask Language Understanding (TMMLU) is curated from examinations in Taiwan, consisting of 55 subjects spanning across multiple disciplines, from vocational to academic fields, and covering elementary to professional proficiency levels. It is designed to identify a model's knowledge and problem-solving blind spots similar to human evaluations. See Appendix \ref{tmmlu_subjects} for the list of subjects.\\

\textbf{Summarization} task requires a model to summarize a given passage in an abstract manner. Extreme Summarization (XSum) dataset evaluates abstractive summarization with 226,711 BBC news articles across diverse domains, aiming for one-sentence summaries. \\
% Earnings Call Transcripts (ECTSum) is a dataset featuring transcripts of public companies' earnings calls and their corresponding short, expert-written bullet point summaries from Reuters articles, highlighting the unstructured and varied nature of the summarization task. 

%\textbf{Scientific Papers} dataset comprises long, structured documents sourced from the ArXiv and PubMed OpenAccess repositories.....

\textbf{Classification} task is defined as requesting a model to determine the category of given input text, such as sentiment analysis and natural language inference. IMDB dataset offers binary sentiment classification with 25,000 polar movie reviews each for training and testing sentiment classifiers. \\

% Amazon Polarity dataset comprises ~35 million reviews from an 18-year span up to March 2013, including product details, user information, ratings, and plaintext reviews for sentiment classification.

% \textbf{Reasoning} requires a model to reason through several steps. GSM8K is a collection of 8.5K high-quality grade school math problems, divided into 7.5K for training and 1K for testing, requiring 2 to 8 arithmetic steps for solutions, suitable for evaluating the reasoning capability of models. \\

% \textbf{Real-world Annotated Few-shot Tasks} (RAFT) dataset, .... 

% sourced from the Yelp Dataset Challenge 2015, includes the Yelp reviews full star dataset with 650,000 training and 50,000 testing samples, distributed evenly across 1 to 5 star ratings.

%\textbf{Rewriting} .... \textbf{ParaSCI} dataset comprises 350,044 paraphrase pairs, with sentences that are notably longer and more divergent than other paraphrase datasets, suitable for evaluating paraphrase generation.

\textbf{Table Understanding} task evaluates a model's capacity to construct an accurate depiction of the data presented to it in both tabular and natural language formats, and its ability to identify and retrieve the pertinent details required to address a straightforward query.
The ``penguins in a table'' task contained in BIG-bench asks a language model to answer questions about the animals contained in a table, or multiple tables, described in the context. \\

%\textbf{Output format generation} .... \textbf{Amazon Review Multi}

% \textbf{Tool Using} task measures a model's capacity to leverage a correct external tool for help instead of fully relying on its auto-regressive outputs. ToolQA is an open-source dataset for evaluating tool-augmented large language models, with questions that often require multiple tool uses. \\ %The dataset is presented in two difficulty levels: Easy and Hard, and 

To assess the capability of the models, we adopt metrics from academic benchmarks like HELM. For evaluations in areas like Contextual QA, World Knowledge, Classification, and Table Understanding, we provide the prefix exact match (EM) scores. In terms of Summarization, ROUGE-2 is reported. 

\subsection{Helpfulness}
To assess language models' ability to provide helpful answers to open-ended questions, we use TAIDE-14 \citep{taide14_2023}. TAIDE-14 consists of 14 different text generation tasks covering 50 topics and includes a total of 140 prompts specifically designed to evaluate Traditional Chinese LLM. These prompts were created by GPT-4 using the provided task, domain, and keywords, and were further validated by human experts. The 14 task types are listed in the following: open-ended generation, classification, question answering, summarization, writing, translation, text analysis, commonsense reasoning, letter writing, extraction, recommendation, sentiment analysis, providing suggestions, and dialogue generation.

\section{Results}
\subsection{Models Compared}

In this study, we analyze the performances of three models: GPT-3.5, Taiwan-LLaMa-v1.0, Model 7-C and Model 7-C-Chat. The version of GPT-3.5 utilized for this comparison is a snapshot titled GPT-3.5-Turbo-0613, dated June 13, 2023. Taiwan-LLaMa-v1.0, on the other hand, is a refined version of the Llama 2 model, configured for Traditional Chinese. It has been pre-trained on a dataset encompassing over 5 billion tokens and further fine-tuned using a rich set of more than 490,000 instruction-response samples.

\subsection{Capabilities Benchmark Results}

\autoref{measured_tasks_performance} illustrates the comparative performance of various models on designated datasets. We carry out all evaluation zero-shot and use greedy decoding for a fair comparison. It is evident that GPT-3.5 predominantly surpasses other models in benchmark tests spanning all assessed capabilities. Both Taiwan-LLaMa-v1.0 and Model 7-C manage to approximate GPT-3.5's performance in limited instances, exhibiting less than a 5\% discrepancy in certain benchmarks. Specifically, Taiwan-LLaMa-v1.0 showcases parallel performance in the IMDB-TC benchmark, whereas Model 7-C is comparable in the DRCD and XSum-TC benchmarks. \\

Notwithstanding, the table understanding task reveals a discernible deficiency in both Taiwan-LLaMa-v1.0 and Model 7-C, with frequent hallucinations evident in numerous samples. Moreover, summarization tasks delineated suboptimal results; even though the models were instructed to condense the context into a single concise sentence, they demonstrated low Rouge-2 scores universally. This underperformance was manifested as over-extended summaries in GPT-3.5 and Model 7-C, and occasional lack of summaries in Taiwan-LLaMa-v1.0. We assessed the XSum dataset and found the presence of summaries incorporating elements not delineated in the original documents, potentially a causal factor in the diminished performance metrics observed in the tasks. \\

% We observed that the quality of the summary answer key in the XSum evaluation dataset is not satisfactory. These generally low evaluation scores may not accurately reflect the capabilities of the language models. An example are shown in \autoref{question_example}

\begin{table}
    \small
    \centering
    \begin{tabular}{|c|c|SSSS|}
    \hline
        \multirow{2}{*}{\textbf{Capability tested}} & \multirow{2}{*}{\textbf{Dataset (metric)}}  & \multicolumn{4}{c|}{\textbf{Models}} \\
        & & \rotatebox{45}{\textbf{GPT-3.5}}  & \rotatebox{45}{\textbf{Taiwan-LLaMa-v1.0}} & \rotatebox{45}{\textbf{Model 7-C}} & \rotatebox{45}{\textbf{Model 7-C-Chat}}\\ \hline
        \multirow{2}{*}{Contextual QA} & DRCD (EM) & 0.771 & 0.719 & 0.761&\\
        & FGC (EM) & 0.48 & 0.33 & 0.41 & \\\hline
        % Contextual QA & DRCD (F1) & 45.76 & & 6.30 \\ \hline
        \multirow{2}{*}{World Knowledge} & TTQA (EM) & 1.00 & 0.59 & 0.86 & 0.56 \\
        & TMMLU (EM) & 0.515 & 0.307 & 0.391 & \\ \hline
        Summarization & XSum-TC (Rouge-2)\footnote{We sub-sampled 5000 samples from the test set of the original XSum dataset for evaluation.} & 0.032 & 0.001 & 0.035 &\\ \hline
        Classification & IMDB-TC (EM) & 0.941 & 0.929 & 0.842 & 0.916\\ \hline
        % Rewriting (Rouge-2) & & & &\\ \hline
        % Reasoning & GSM8K-TC (EM) & 0.00* & 0.00* & 0.00*\\ \hline
        Table Understanding & Penguins-in-a-Table-TC (EM)\footnote{Although some of the questions in this dataset have their options listed out, and can be considered as multiple choice, the model's output does not always fall into one of the choices, leading to a score lower than random guessing. We include "the ability to identify this task as a multiple choice" as part of the task objective and do not fix this "bug".} & 0.32 & 0.00 & 0.01 & 0.08 \\ \hline
        % Tool Using & ToolQA-TC (Success rate) & & & \\ \hline
        % Output format generation (EM) & & & & \\ \hline
        % Literature-writing (MOS) & & & &\\ \hline
        % Role-play (MOS) & & & & \\ \hline
    \end{tabular}
    \caption{The benchmark result of models.}
    \label{measured_tasks_performance}
\end{table}

%\textbf{Discussion} We observed the summary of XSum is not really 

\subsection{Helpfulness Benchmark Results}

\begin{figure*}[ht]
\centering
\includegraphics[width=1.0\textwidth]{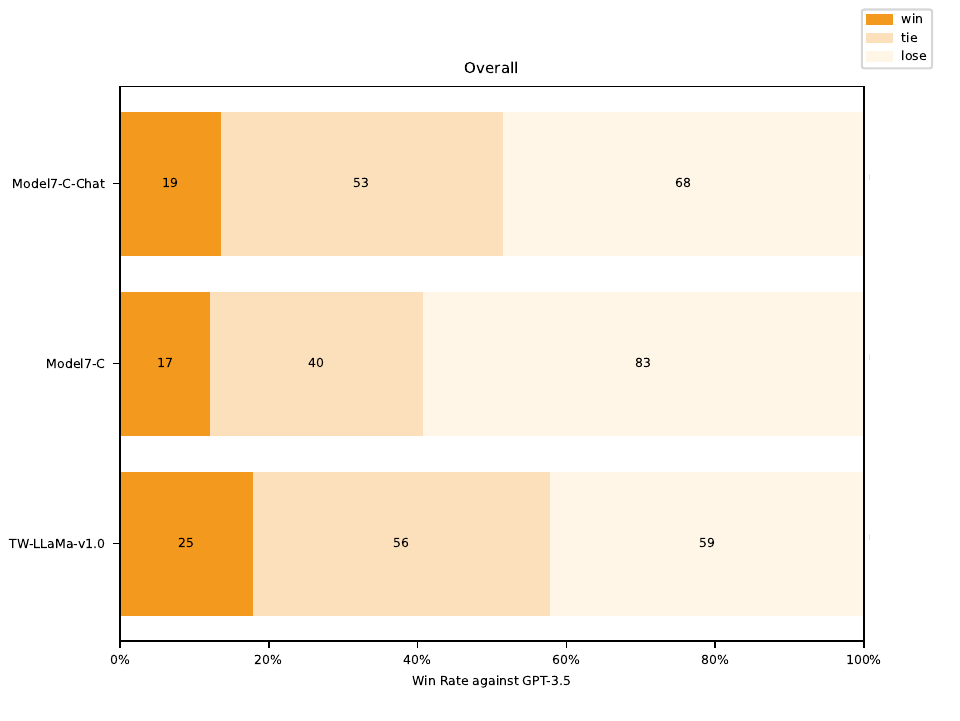}
\caption{Win-rate between models on the TAIDE-14 benchmark. The win-rate chart shows comparisons between GPT-3.5, Taiwan-LLaMa-v1.0, Model 7-C and Model 7-C-Chat.}
   \label{fig:win_rate_comp}
\end{figure*}

We present the win-rate chart to demonstrate the helpfulness of the language models on TAIDE-14 tasks, judged by GPT-4 on a scale of 1 to 6. Our proprietary model fine-tuned for chatting capability, Model7-C-Chat, outperforms GPT-3.5 on 19, and matches GPT-3.5 on 53, out of all the 140 test samples. Though, regarding TAIDE-14, Taiwan-LLaMa-v1.0 shows better capability than our Model 7-C series. See \autoref{fig:win_rate_comp} for reference.

\section{Conclusion}
In conclusion, the evaluation of large language models, particularly in the context of Traditional Chinese, is a critical and challenging task. This study proposes a comprehensive set of benchmarks, built upon existing Traditional Chinese and English datasets, to assess the capabilities of these models across various tasks.  The evaluation of models such as GPT-3.5, Taiwan-LLaMa-v1.0, and our proprietary model, Model 7-C, demonstrated the effectiveness of these benchmarks. Notably, Model 7-C showed comparable performance to the state-of-the-art GPT-3.5 model regarding certain of evaluated Traditional Chinese tasks. \\

The introduction of these benchmarks is a significant step towards advancing the evaluation of language models in Traditional Chinese. By making our benchmark code and relevant datasets open-source, and releasing our base model, Model 7-C, for trial, we aim to stimulate further research in this field. We believe that these resources will provide a valuable foundation for future studies aiming to improve the capabilities of language models in Traditional Chinese.

%\section{Acknowledgement}
% About people who've helped the release of this work
% Acknowledge TAIDE and test TAIDE

% \section{References}
% Reference goes here

\begin{CJK*}{UTF8}{bsmi}
\bibliographystyle{plainnat}
\bibliography{reference}

\appendix
\section{Benchmark Question Examples}
\label{question_example}
\begin{CJK*}{UTF8}{bsmi}
\begin{itemize}

\item DRCD
\begin{mdframed}
要探討從梨俱吠陀到波你尼時代梵語的發展，可以考察印度教其它文本，如娑摩吠陀、夜柔吠陀、阿闥婆吠陀、梵書和奧義書。在此期間，這門語言的威望、它的神聖用途及其正確發音的重要性，形成了一股強大的保守力量，防止梵語像普通語言一樣隨時間而演變。現存最古老的梵語文法是波你尼的《八篇書》，大約於公元前四世紀成形。它本質上是規範性文法，就是說它定義了正確梵語的用法，儘管它包含了描述成分，但大多是處理在波你尼時代已經廢棄了的某些吠陀形式。這裡所說的「梵語」不作脫離於其他語言的特殊語言看待，而是視作講話的高雅純正或完美方式。通過梵語文法家如波你尼的精密分析，梵語的知識在古印度是社會等級層次高和教育程度高的標誌，並主要教授給高等世襲階級的成員。梵語作為古印度的學術語言，與俗語同時共存，而俗語演化成了中古印度-雅利安語方言，並最終演化成了當代的各種印度-雅利安語言。
\newline\newline Q: 夜柔吠陀與阿闥婆吠陀均可以最為研究哪一門語言的參考？
\newline A: 梵語
\end{mdframed}

\item TTQA
\begin{mdframed}[innerleftmargin=+10pt]
它是位於亞熱帶的台灣內唯一一種溫帶性魚類，也是只產於台灣的特有櫻鮭亞種，為冰河孑遺生物。由於其相當稀有且瀕臨絕種，加上它的生活習性迥異於其他魚類，遂得「國寶魚」之美譽。
\newline\newline Q: 該動物的名稱是：
\newline A: 櫻花鉤吻鮭
\end{mdframed}

\item FGC
\begin{mdframed}
海倫·凱勒於1880年6月27日出生在美國阿拉巴馬州的塔斯坎比亞 。海倫·凱勒原為健康的嬰兒，但在19個月大的時候患了急性腦充血病，失去了聽覺和視覺。長大後運用自創的手語與家庭成員溝通。隨著年歲的增長，簡單的交流不能滿足她，脾氣變得暴躁。6歲時，她的父母在家庭醫生的協助下，邀請柏金斯啟明學校的安妮·蘇利文老師作為海倫·凱勒的啟蒙導師。在1887年，藉著她的導師安妮·蘇利文對她耐心的教導和關愛，並找到專家使她學會發音，讓她學會流暢的表達，才開始與其他人溝通並接受教育。海倫·凱勒不但學會閱讀和說話，還以驚人的毅力完成了哈佛大學的學業並於1904年畢業，成為有史以來第一個獲得文學學士學位的盲聾人士。成年後，她繼續廣泛閱讀刻苦學習，掌握了英語、法語、德語、拉丁語和希臘語，成為盲聾的作家和教育家。她致力於殘疾人事業，四處募捐以改善殘疾人的生活環境和受教育水平。她的事跡使她入選美國《時代周刊》「人類十大偶像之一」，被授予「總統自由獎章」。
\newline\newline Q: 海倫凱勒出生於哪一個城市？
\newline A: 塔斯坎比亞
\end{mdframed}

\item TMMLU
\begin{mdframed}
Q: 「臺灣原住民的布只有形制屬傳統或較現代的分別，像圓領的剪裁、鈕扣和棉布的使用等，都是受漢人的影響而來。泰雅族的貝珠鈴衣，是貝珠串底下加銅鈴裝飾，銅鈴也是和漢人交易而來。日治時代的原住民服裝，還出現以漢人棉布做底、日本布做袖口、原住民圖案做主要裝飾的混搭法。」這段文字的主旨最可能是下列何者？
\newline(A)不同文化的碰撞，可融合並產生新的火花
\newline(B)外來文化的入侵，讓在地的傳統文化日漸消失
\newline(C)臺灣原住民的文化，影響了漢人與日本人的穿著
\newline(D)觀察不同族群的服飾，就能了解不同文化的差異
\newline\newline A: (A)
\end{mdframed}

\item XSum-TC
\begin{mdframed}
埃文斯一開始就以一記漂亮的弧線球從 20 碼處射入底角，幫助矮腳雞隊取得領先。蝦米隊發起反擊，瑞恩·倫納德 (Ryan Leonard) 的一記猛烈遠射迫使布拉德福德門將本·威廉姆斯 (Ben Williams) 做出精彩撲救。當蒂龍·巴內特的凌空抽射擊中橫梁時，主隊幾乎扳平比分，但布拉德福德堅持了下來。
\newline\newline Summary: 憑藉李·埃文斯的早早進球，布拉德福德城擊敗紹森德聯隊，確保獲得聯賽附加賽席位。
\end{mdframed}

\item IMDB-TC
\begin{mdframed}
我不在乎是否有人認為這部電影不好。如果你想知道真相，這是一部非常好的電影！它具有電影應有的一切。你真的應該買這個。
\newline\newline Sentiment: 正面
\end{mdframed}

% \item TAIDE-14
% \begin{mdframed}
% Q: 請將以下生物技術應用分為低風險、中風險、和高風險三種類型，並簡要解釋為什麼將其分為該風險類型。
% \newline基因編輯、生物染料、基因療法、基因工程作物、細胞再生、克隆技術、人類胚胎研究、生物能源。
% \newline\newline A: \\
% 低風險：  \\
% 1. 生物染料：生物染料通常是使用微生物或植物來生產的，對環境和人類健康的影響相對較小。\\
% 2. 生物能源：生物能源是利用生物質或生物廢料來生產能源，通常被認為是可再生和環保的能源。\\
% \\
% 中風險：\\
% 1. 基因編輯：基因編輯技術可以在生物體內進行精確的基因修改，但可能對生態系統產生未知的影響。\\
% 2. 基因工程作物：基因工程作物具有抗蟲害和耐草甘膦等特性，有利於農業生產，但可能對生態系統產生潛在影響。\\
% 3. 細胞再生：細胞再生技術可以幫助治療一些疾病，但仍然存在一定程度的技術風險和道德爭議。\\

% 高風險：\\
% 1. 基因療法：基因療法目前仍處於初級階段，可能帶來不可預知的副作用和安全風險。\\
% 2. 克隆技術：克隆技術涉及復制生物體，可能對生物多樣性和生態平衡產生影響。\\
% 3. 人類胚胎研究：人類胚胎研究涉及胚胎的操縱和修改，具有較高的道德和生命安全風險。\\
% \end{mdframed}

\end{itemize}
\end{CJK*}

\section{Taiwan Massive Multitask Language Understanding (TMMLU)}
\label{tmmlu_subjects}

%\textbf{BELOW IS A TABLE OF SUBJECTS IN TMMLU}
\centering
\begin{tabular}{|l|ccc|}
\hline
\textbf{Subject} & \textbf{Model 7-C} & \textbf{Taiwan-LlaMa-v1.0} & \textbf{GPT-3.5} \\
\hline
企業管理 & 32/50 & 23/50 & 37/50 \\
國際關係與近代外交史 & 9/20 & 4/20 & 10/20 \\
基礎醫學 & 38/76 & 25/76 & 59/76 \\
分科測驗物理 & 0/9 & 2/9 & 0/9 \\
中式麵食加工 & 53/119 & 37/119 & 71/119 \\
普通物理 & 10/30 & 8/30 & 13/30 \\
分科測驗地理 & 12/25 & 6/25 & 13/25 \\
油壓 & 18/47 & 23/47 & 26/47 \\
分科測驗歷史 & 21/38 & 15/38 & 23/38 \\
國際法 & 11/21 & 4/21 & 14/21 \\
中醫臨床醫學 & 97/325 & 77/325 & 133/325 \\
中餐烹調─葷食 & 32/60 & 22/60 & 32/60 \\
中餐烹調─素食 & 27/59 & 22/59 & 35/59 \\
普通生物學 & 23/48 & 16/48 & 36/48 \\
會考國文 & 24/63 & 24/63 & 33/63 \\
冷凍空調 & 16/56 & 17/56 & 25/56 \\
分科測驗公民與社會 & 9/27 & 9/27 & 14/27 \\
分科測驗化學 & 2/7 & 1/7 & 1/7 \\
機電整合 & 20/44 & 17/44 & 23/44 \\
分科測驗數學甲 & 2/5 & 1/5 & 0/5 \\
法學知識 & 14/30 & 13/30 & 25/30 \\
營養學 & 33/86 & 23/86 & 39/86 \\
分科測驗生物 & 6/21 & 5/21 & 8/21 \\
商業道德 & 29/80 & 26/80 & 23/80 \\
用電設備檢驗 & 21/57 & 14/57 & 31/57 \\
平版印刷 & 25/60 & 26/60 & 28/60 \\
建築塗裝 & 31/60 & 24/60 & 38/60 \\
普通化學 & 15/52 & 17/52 & 22/47 \\
行銷管理 & 20/51 & 14/51 & 28/51 \\
會考數學 & 2/7 & 2/7 & 2/7 \\
會計學經濟學 & 15/41 & 11/41 & 12/41 \\
資訊安全 & 35/75 & 23/75 & 48/75 \\
農田灌溉排水─灌溉水質管理及檢驗項 & 26/55 & 21/55 & 29/55 \\
牙醫學 & 159/454 & 135/454 & 214/454 \\
社會工作大意 & 26/50 & 19/50 & 32/50 \\
配電電纜裝修 & 20/56 & 19/56 & 26/56 \\
綜合法政知識 & 28/57 & 18/57 & 37/57 \\
網路架設 & 26/58 & 19/58 & 44/58 \\
飛機修護 & 27/60 & 17/60 & 41/60 \\
總體經濟 & 12/45 & 10/45 & 19/45 \\
臨床心理學 & 110/259 & 86/259 & 155/259 \\
臨床血清免疫學與臨床病毒學 & 32/78 & 22/78 & 52/78 \\
裝潢木工 & 22/58 & 15/58 & 32/58 \\
製鞋─製配底 & 27/57 & 16/57 & 34/57 \\
製鞋─製面 & 19/59 & 13/59 & 26/59 \\
觀光資源概要 & 24/48 & 22/48 & 31/48 \\
計算機數學 & 2/10 & 1/10 & 0/10 \\
車輛塗裝 & 23/59 & 21/59 & 34/59 \\
通信技術 & 24/54 & 16/54 & 24/54 \\
邏輯推理 & 1/15 & 4/15 & 9/15 \\
醫學 & 14/27 & 13/27 & 19/27 \\
金銀珠寶飾品加工 & 21/60 & 25/60 & 33/60 \\
電路電子學 & 5/13 & 3/13 & 10/13 \\
食品檢驗分析 & 29/60 & 24/60 & 40/60 \\
\hline
\end{tabular}

\end{CJK*}

\end{document}